\def\year{2022}\relax
\documentclass[letterpaper]{article} 
\usepackage{aaai22}  
\usepackage{times}  
\usepackage{helvet}  
\usepackage{courier}  
\usepackage[hyphens]{url}  
\usepackage{graphicx} 
\usepackage{natbib}  
\usepackage{caption} 
\usepackage{color}
\usepackage{siunitx}
\usepackage{bm}
\usepackage{multirow}
\usepackage{dsfont}
\usepackage{tabularx}
\usepackage{caption}
\usepackage{verbatim}
\usepackage{float}
\usepackage{pifont}
\usepackage{amsthm}
\usepackage{verbatim}
\usepackage{rotating} 
\usepackage{tabularx}
\usepackage{amsmath}
\usepackage{amssymb}
\usepackage{booktabs}
\usepackage{mathrsfs}
\usepackage{float}
\DeclareCaptionStyle{ruled}{labelfont=normalfont,labelsep=colon,strut=off} 
\usepackage{algorithm}
\usepackage{algorithmic}

\usepackage{newfloat}
\usepackage{listings}
\floatstyle{ruled}
\newfloat{listing}{tb}{lst}{}
\floatname{listing}{Listing}
\title{Domain Disentangled Generative Adversarial Network for Zero-Shot Sketch-Based 3D Shape Retrieval}
\author{
	Rui Xu,
	Zongyan Han,
	Le Hui,
	Jianjun Qian,
	Jin Xie\thanks{Corresponding author.}
}
\affiliations{
	PCA Lab, Key Lab of Intelligent Perception and Systems for High-Dimensional Information of Ministry of Education\\
	Jiangsu Key Lab of Image and Video Understanding for Social Security\\
	School of Computer Science and Engineering, Nanjing University of Science and Technology, Nanjing, China\\
	\{xu\_ray, hanzy, le.hui, csjqian, csjxie\}@njust.edu.cn
}

\usepackage{bibentry}
\begin{document}

\maketitle

\begin{abstract}
Sketch-based 3D shape retrieval is a challenging task due to the large domain discrepancy between sketches and 3D shapes. Since existing methods are trained and evaluated on the same categories, they cannot effectively recognize the categories that have not been used during training. In this paper, we propose a novel domain disentangled generative adversarial network (DD-GAN) for zero-shot sketch-based 3D retrieval, which can retrieve the unseen categories that are not accessed during training. Specifically, we first generate domain-invariant features and domain-specific features by disentangling the learned features of sketches and 3D shapes, where the domain-invariant features are used to align with the corresponding word embeddings. Then, we develop a generative adversarial network that combines the domain-specific features of the seen categories with the aligned domain-invariant features to synthesize samples, where the synthesized samples of the unseen categories are generated by using the corresponding word embeddings. Finally, we use the synthesized samples of the unseen categories combined with the real samples of the seen categories to train the network for retrieval, so that the unseen categories can be recognized. 
In order to reduce the domain shift problem, we utilized unlabeled unseen samples to enhance the discrimination ability of the discriminator. With the discriminator distinguishing the generated samples from the unlabeled unseen samples, the generator can generate more realistic unseen samples.
Extensive experiments on the SHREC'13 and SHREC'14 datasets show that our method significantly improves the retrieval performance of the unseen categories.
\end{abstract}

\section{Introduction}

In recent years, with the massive increase of 3D models, 3D shape retrieval has attracted widespread attention. Existing methods mainly contain three categories, including shape-based methods~\cite{iyer2005three,xie2015deepshape,zhu2016deep,jiang2019mlvcnn}, text-based methods~\cite{min2004comparison,goldfeder2008autotagging} and sketch-based methods~\cite{eitz2012sketch,furuya2013ranking,li2014comparison,dai2018deep}.Compared with text and 3D shape, sketch is more convenient and easier to obtain. However, due to the abstraction of sketches and the discrepancy between 2D sketches and 3D shapes, sketch-based 3D shape retrieval is still a challenging problem.

Recently, many efforts have been made in sketch-based 3D retrieval. In the early years, many works improve the performance of sketch-based 3D retrieval by learning robust 3D shape features \cite{wang2015sketch,xie2017learning,zhu2016heat,zhu2016learning,xu2020sketch}.
\cite{wang2015sketch} selected two different views with angles larger than 45 degrees to characterize 3D shapes.
And a siamese convolutional neural network was used to extract the features of sketches and projections of 3D shapes.
To further improve 3D shape representations, \cite{xie2017learning} proposed to learn the Wasserstein barycenters of the multi-views projections.
In addition, how to measure the cross-domain similarity between sketches and 3D shapes is also crucial \cite{dai2017deep,he2018triplet}.
In \cite{dai2017deep}, the discriminative loss is proposed to increase the distinction of different categories in each domain, and the correlation loss is used to minimize the domain discrepancy between sketch and 3D shape.
In sketch-based 3D shape retrieval, these methods have achieved remarkable results in categories that are used for both training and evaluation. 
Nonetheless, these traditional methods cannot effectively retrieve the unseen categories that are not used during training.

In this paper, we propose a novel domain disentangled generative adversarial network (DD-GAN) to effectively retrieve the unseen categories in sketch-based 3D shape retrieval. The key idea of our method is to utilize the generative adversarial network to generate samples of the unseen categories. After obtaining these unseen samples, we can convert the zero-shot learning task into a traditional supervised learning task. Specifically, in our DD-GAN, we first use SketchCNN and ShapeCNN to extract features of sketches and 3D shapes, respectively. We disentangle the features of each domain by using three fully connected layers to obtain domain-invariant features ($i.e.$, pure semantic information) and domain-specific features ($i.e.$, contour style or texture information). Then, we propose the semantic alignment module to enhance the discrimination of the domain-invariant features between different categories and align the domain-invariant features with the corresponding word embedding of the category.
In the semantic alignment module, we adopt the triplet loss to minimize the intra-class distance of domain-invariant features and maximize the inter-class distance of domain-invariant features.
In addition, we align the domain-invariant features with the corresponding word embedding by computing the cosine similarity between them.
After that, the domain-specific features are used as the condition combined with the domain-invariant features to generate samples of the corresponding categories by the generative adversarial network. 
We employed an efficient adversarial loss to train our generative model to make the generated samples more realistic.

In order to alleviate the domain shift problem of the unseen categories, we use the word embedding of the unseen categories combined with domain-specific features of the seen categories to generate the samples of the unseen categories. By using the discriminator to distinguish the generated samples and the real unseen samples, we can reduce the domain shift of the unseen categories. We use the generated samples of the unseen categories combined with the real samples of the seen categories to train our network. We use the obtained domain-invariant features for retrieval. Experimental results on the SHREC'13 and SHREC'14 datasets demonstrate the effectiveness of the proposed method for sketch-based 3D shape retrieval. Especially, our method can effectively retrieve the unseen categories.

In summary, our main contributions are as follows: (1) To the best of our knowledge, we are the first to consider zero-shot sketch-based 3D shape retrieval. (2) We propose a novel domain disentangled generative adversarial network (DD-GAN) that can learn the discriminative features of different domains by decomposing and combining domain-invariant features and domain-specific features to generate samples of different categories. 
(3) We extend our DD-GAN to the transductive setting by utilizing unlabeled unseen samples through our generative model. With the discriminator distinguishing the generated samples from the unlabeled unseen samples, the generator can generate more robust unseen samples to alleviate the domain shift problem.
(4) Experimental results on the SHREC'13 and SHREC'14 datasets show that our method can effectively retrieve the unseen categories.

\section{Related Work}

\subsection{Sketch-Based 3D Shape Retrieval}
In the decades, researches employed various hand-crafted features to describe sketches and 3D shapes \cite{saavedra2012sketch,li2013sketch,li2017sketch,yoon2017sketch}.
\cite{yoon2017sketch} proposed a sparse coding based methods to match the HOG-SIFT features of sketches and 3D objects.
\cite{saavedra2012sketch} used histogram of keyshape orientations (HKO) as global descriptors to determine the appropriate viewpoint of 3D shapes, then employed keyshape angular spatial distribution (KASD) as local descriptors to match the sketches and 3D shapes. 
\cite{li2017sketch} proposed a viewpoint entropy distribution to describe 3D shapes, and obtained a set of representative sample views of 3D shapes by adaptive view clustering for 2D-3D comparison.

Recently, with the rapid development of deep learning, deep features extracted by neural networks gradually replaces traditional hand-crafted features \cite{tasse2016shape2vec,kuwabara2019query,chen2019deep,dai2020cross,liu2021guidance}.
Many methods focus on how to extract robust sketch features and 3D shape features, and effectively measure the similarity between the sketch and 3D shape by reducing domain gap, so as to achieve cross-domain retrieval.
\cite{xu2020sketch} selected the best perspective projections of the 3D shapes according to the perspective of the training sketches, and used MVCNN \cite{su2015multi} to extract the features of these projections to obtain the robust 3D shape features.
Based on the multi-view pairwise relationship(MVPR) learning, \cite{li2017multi} proposed a probabilistic framework to infer the pairwise relationship between sketches and projections of 3D shapes to tackle the retrieval problem.
\cite{chen2018deep} designed a transformation network to transform sketch features into 3D shape feature space. At the same time, they used the cross-modality mean discrepancy minimization to enhance the correlations of transformed sketch features and 3D shape features.
\cite{he2018triplet} proposed a novel triplet-center loss, which can directly minimize the intra-class distance and maximize the inter-class distance for the two different domains.
To alleviate the domain gap between sketch and 3D shape, \cite{qi2018semantic} proposed a novel method to map sketches and 3D shapes into a joint semantic embedding space.

\subsection{Zero-Shot Learning}
Nowadays, zero-shot learning has attracted widespread attention and has been applied to various visual tasks, such as image classification \cite{xian2016latent,mensink2014costa,bucher2017generating,han2020learning}, semantic segmentation \cite{bucher2019zero,kato2019zero} and sketch-based image retrieval (SBIR) \cite{dey2019doodle,dutta2019style,pandey2020stacked,xu2020progressive}.
Many methods map visual features to high-dimensional semantic space, and utilize attributes as a ``bridge" for knowledge transfer from seen categories to unseen categories \cite{akata2015label,kodirov2017semantic,han2020learning,han2021contrastive}.
In addition, there are other methods that use generative models to generate unseen classes and train their models with the generated unseen categories, so as to improve the robustness of models to real unseen categories \cite{xian2019f,huang2019generative,gao2020zero}.
For more information about the zero-shot topic, please refer to the comprehensive survey\cite{xian2018zero}. 

Actually, our task is related to zero-shot sketch-based image retrieval (SBIR), which involves two different domains.
\cite{yelamarthi2018zero} used conditional variational autoencoder (CVAE) and adversarial autoencoder (CAAE) to associate the characters of sketch with that of the image.
\cite{dutta2019semantically} proposed a semantically aligned paired cycle-consistent generative (SEM-PCYC) model to map the visual features of sketch and image to a common semantic space. During the testing phase, they used the learned mappings to generate embeddings of unseen classes for retrieval.
\cite{dutta2019style} designed a style-guide image generator to generate fake images from sketch, and retrieved images through generated images to obtain the final retrieval results.
To prevent the catastrophic forgetting phenomenon, \cite{liu2019semantic} proposed semantic-aware knowledge preservation (SAKE) to preserve previously acquired knowledge during fine-tuning.
\cite{deng2020progressive} proposed a progressive cross-domain semantic network to solve the knowledge loss problem. In addition, they formulated a cross-reconstruction loss to reduce the domain gap.

\section{Methodology}

Given the dataset $\mathcal{D}=\{(\bm{x}_{i},\bm{y}_{i},c_{i})\mid c_i \in \mathcal{C}\} $, where $\bm{x}_i$, $\bm{y}_i$, and $c_i$ are the sketch, 3D shape, and the label of the $i$-th sample, respectively. Here, $\mathcal{C}$ denotes the set of different categories. In the zero-shot setting, we split the whole categories into two sets $\mathcal{C}_{seen}$ and $\mathcal{C}_{unseen}$, where $\mathcal{C}_{seen} \cup \mathcal{C}_{unseen} = \mathcal{C} $ and $ \mathcal{C}_{seen} \cap \mathcal{C}_{unseen} = \varnothing$. According to the sets $\mathcal{C}_{seen}$ and $\mathcal{C}_{unseen}$, we can obtain the  training set $\mathcal{D}_{train} = \{(\bm{x}_{i},\bm{y}_{i},c_{i})\mid c_{i} \in \mathcal{C}_{seen} \} $ and test set $ \mathcal{D}_{test} = \{(\bm{x}_{i},\bm{y}_{i},c_{i})\mid c_i \in \mathcal{C}_{unseen}\}$. In addition, zero-shot learning (ZSL) can be divided into the inductive setting and the transductive setting according to the availability of unlabeled data. In the inductive setting, we use the set $\mathcal{D}_{train}$ to train the network and use the set $\mathcal{D}_{test}$ for evaluation. However, in the transductive setting, in addition to using the set $\mathcal{D}_{train}$ to train the network, we additionally use the set $\mathcal{D}_{test}$, but do not use the ground truth. Note that in this paper, our method is under the transductive setting.

\subsection{Domain Disentangling}

\begin{figure*}[t]
	\begin{center}
		\includegraphics[width=1.0\linewidth]{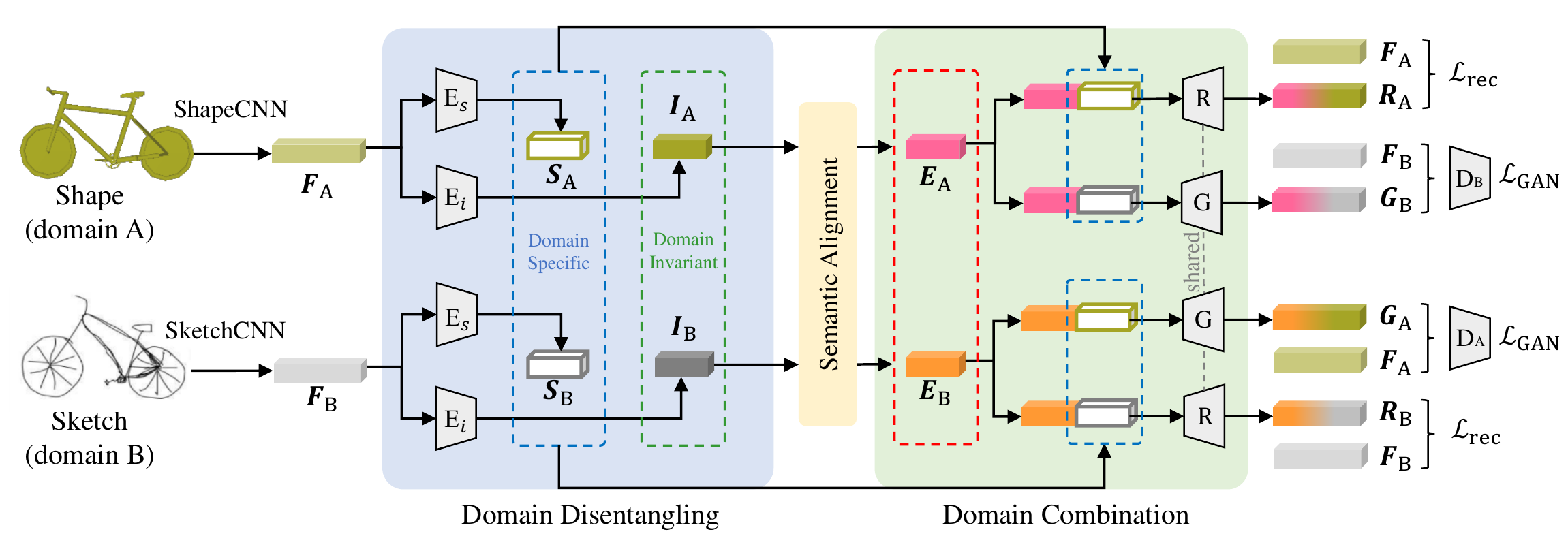}
	\end{center}
	\vskip -10pt
	\caption{Illustration of our domain disentangled generative adversarial network (DD-GAN). Specifically, we use feature disentanglement to obtain domain-specific features and domain-invariant features from sketch features and 3D shape features. The semantic alignment module is used to enhance the discrimination of domain-invariant features, and align semantic features mapped from visual features with corresponding word embeddings. The domain combination module completes feature reconstruction and cross-domain generation of sketches and 3D shapes by combining semantic features with different domain-specific features.}
	\label{fig:framework}
	\vskip -10pt
\end{figure*}

\textbf{Feature Disentanglement.} In order to reduce the domain discrepancy between 2D sketches and 3D shapes, we develop the domain disentangling module to extract the domain-invariant features and domain-specific features. As shown in Figure~\ref{fig:framework}, we depict the details of domain disentangling. Given the samples of the 3D shape (dubbed domain A) and sketch (dubbed domain B), we first use ShapeCNN and SketchCNN to extract the initial features $\bm{F}_A\in\mathbb{R}^{D}$ and $\bm{F}_B\in\mathbb{R}^{D}$, where $D$ is the dimension of the feature vector. Based on multi-view convolution neural network (MVCNN)~\cite{su2015multi}, ShapeCNN uses the pre-trained ResNet-50~\cite{he2016deep} on the multiple 2D images to extract initial feature $\bm{F}_A$. SketchCNN also uses the pre-trained ResNet-50 on the sketch to extract initial features $\bm{F}_B$.

Once we obtain the features $\bm{F}_A$ and $\bm{F}_B$ from two domains, we employ the domain-invariant encoder $E_i$ and domain-specific encoder $E_s$ to obtain domain-invariant and domain-specific features, respectively. The domain-invariant features and domain-specific features of the 3D shape and sketch are formulated as:
\begin{equation}
	\begin{aligned}
		\bm{I}_{A} = E_i(\bm{F}_{A}), \bm{I}_{B} = E_i(\bm{F}_{B}) \\
		\bm{S}_{A} = E_s(\bm{F}_{A}), \bm{S}_{B} = E_s(\bm{F}_{B})
	\end{aligned}
\end{equation}
where $\bm{I}_{A}\in\mathbb{R}^{D'}$, $\bm{I}_{B}\in\mathbb{R}^{D'}$, $\bm{S}_{A}\in\mathbb{R}^{D'}$, and $\bm{S}_{B}\in\mathbb{R}^{D'}$ are the domain-invariant features and domain-specific features of the sketch and 3D shape, respectively. It is desired that the domain-invariant features can characterize the semantic information, while the domain-specific features can preserve the unique characteristic (such as texture information) of the domain itself. In the experiment, we use three fully connected layers to implement the encoders $E_i$ and $E_s$. Note that the encoders are not shared in different domains.

\begin{figure}[htb]
	\centering
	\includegraphics[width=0.98\linewidth]{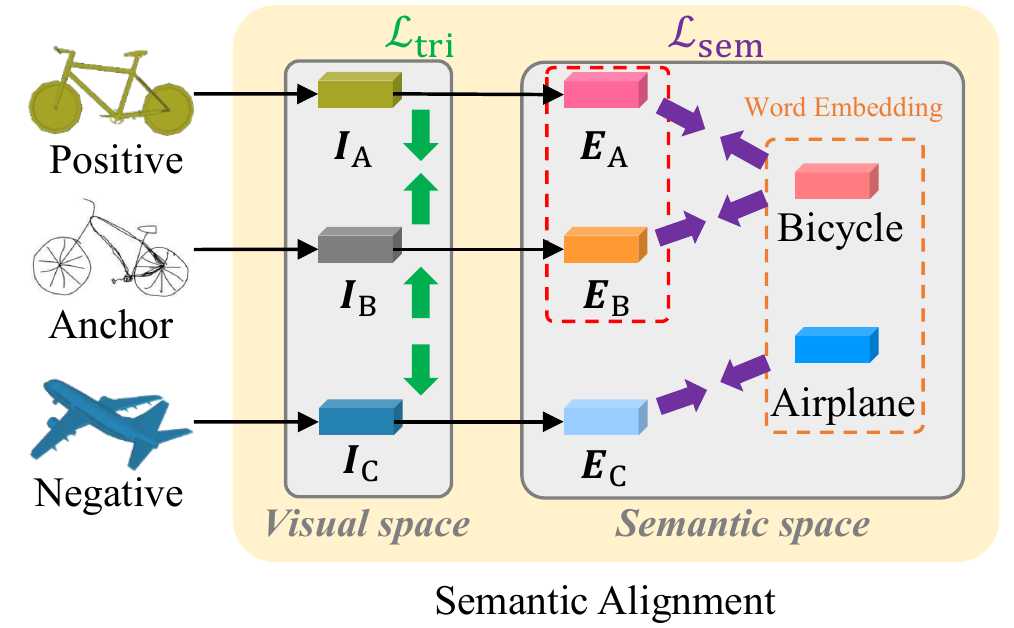}
	\caption{Semantic alignment module.}
	\label{fig:semantic_alignment}
\end{figure}

\textbf{Semantic alignment.} In order to identify the unseen categories, we develop the semantic alignment module by mapping the domain-invariant features to the word embedding space to transfer the knowledge from the seen categories to the unseen categories. 
Here, we also employ triplet loss to enhance the discrimination of domain-invariant features of different categories in the visual space.
As shown in Figure~\ref{fig:semantic_alignment}, given the triplet of anchor, positive, and negative, we first obtain the domain-invariant features $\bm{I}_A$, $\bm{I}_B$, and $\bm{I}_C$, respectively. Then, we perform the triplet loss to minimize the distance from the anchor sample ($\bm{I}_B$) to the positive sample ($\bm{I}_A$) and maximize the distance from the anchor sample ($\bm{I}_B$) to the negative sample ($\bm{I}_C$). 
After that, we use the fully connected layers to align domain-invariant features with word embeddings, which is formulated as:

\begin{equation}
	\begin{aligned}
		\bm{E}_{A} = \phi(\bm{I}_{A}), \bm{E}_{B} = \phi(\bm{I}_{B}), \bm{E}_{C} = \phi(\bm{I}_{C})
	\end{aligned}
\end{equation}
where the $\bm{E}_{A}\in\mathbb{R}^{d}$, $\bm{E}_{B}\in\mathbb{R}^{d}$ and $\bm{E}_{C}\in\mathbb{R}^{d}$ are the semantic features of $\bm{A}$, $\bm{B}$, and $\bm{C}$, respectively. $\phi(\cdot)$ is implemented by the three-layer fully connected layers. Finally, given the word embedding $W=\{\bm{w}_i\in\mathbb{R}^{d} \mid i=1,\ldots,\|\mathcal{C}\|\}$, where $\mathcal{C}$ indicates the set of categories and $d$ is the dimension of the word embedding. By minimizing the cosine distance from the domain-invariant features to the word embedding of the corresponding category, we can align the semantic space to the word embedding space. Therefore, we can recognize the unseen category by using the corresponding word embedding to characterize the semantic information of the unseen category.

\subsection{Domain Combination}

\begin{figure}[htb]
	\centering
	\includegraphics[width=1.0\linewidth]{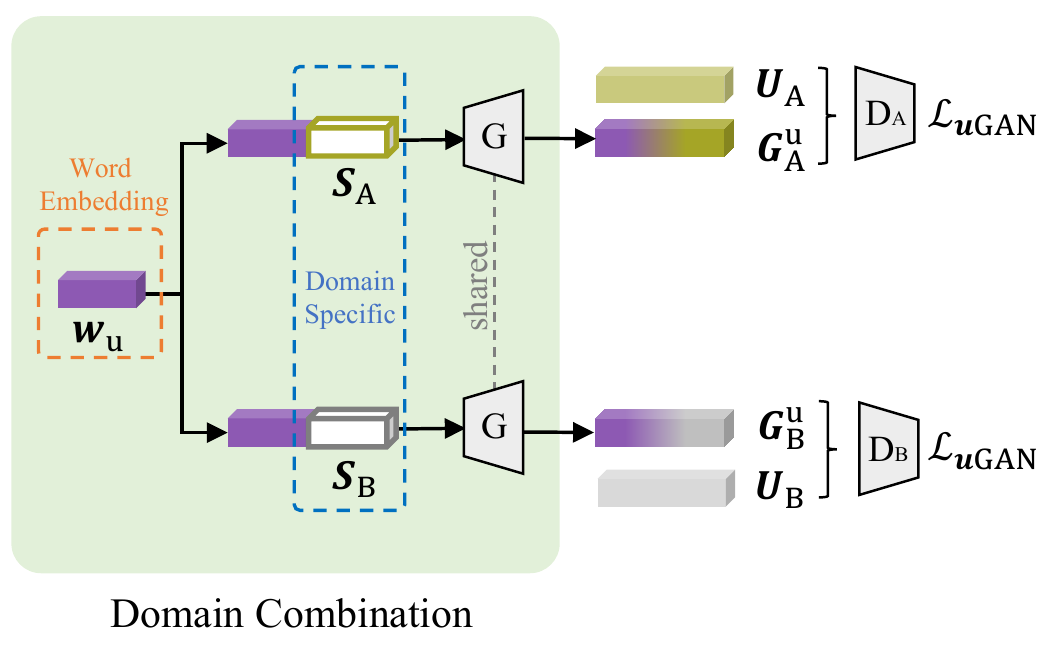}
	\caption{The generation of unseen samples. $\bm{U}_A$ and $\bm{U}_B$ are the initial features of unlabeled 3D shape and sketch.}
	\label{fig:transductive}
\end{figure}

After semantic alignment, we develop the domain combination module by combining the aligned domain-invariant features and domain-specific features to reconstruct initial features and generate samples of different domains at the same time. As shown in Figure~\ref{fig:transductive}, we illustrate the detailed structure of the domain combination module. Specifically, given the aligned domain-invariant features $\bm{E}_A$ and $\bm{E}_B$ of the 3D shape and sketch, we concatenate the aligned domain-invariant features with the corresponding domain-specific features to reconstruct the initial feature, which is written as:
\begin{equation}
	\begin{aligned}
		\bm{R}_{A} = \mathcal{R}([\bm{E}_{A};\bm{S}_{A}]), \bm{R}_{B} = \mathcal{R}([\bm{E}_{B};\bm{S}_{B}])
	\end{aligned}
\end{equation}
where $\bm{R}_{A}\in\mathbb{R}^{D}$ and $\bm{R}_{B}\in\mathbb{R}^{D}$ are the reconstructed features of the 3D shape and sketch, respectively. $[\cdot;\cdot]$ is the concatenation operation. We perform the $L_1$ loss to minimize the distance from the reconstructed features ($\bm{R}_{A}$, $\bm{R}_{B}$) to the corresponding features ($\bm{F}_{A}$, $\bm{F}_{B}$). In addition, we combine the domain-invariant feature with another domain-specific feature to generate the sample of another domain, which is formulated as:
\begin{equation}
	\begin{aligned}
		\bm{G}_{A} = \mathcal{G}([\bm{E}_{B};\bm{S}_{A}]), 
		\bm{G}_{B} = \mathcal{G}([\bm{E}_{A};\bm{S}_{B}]) \end{aligned}
\end{equation}
where $\bm{G}_{A}\in\mathbb{R}^{D}$ and $\bm{G}_{B}\in\mathbb{R}^{D}$ are the generated features of the 3D shape and sketch, respectively. After that, we formulate the adversarial loss by feeding the original feature ($\bm{F}_{A}$, $\bm{F}_{B}$) and the generated features ($\bm{G}_{A}$, $\bm{G}_{B}$) to the discriminators ($D_A$, $D_B$).

\textbf{Transductive setting.} In order to reduce the domain shift between the generated unseen categories and the real unseen categories, we extend our DD-GAN to a transductive setting. As shown in Figure~\ref{fig:transductive}, we illustrate the details of the transductive setting. Given the 3D shape and sketch of the unseen categories, we first use the ShapeCNN and SketchCNN to extract the initial features $\bm{U}_A\in\mathbb{R}^D$ and $\bm{U}_B\in\mathbb{R}^D$. Then, by combining the word embedding $\bm{w}_{u}$ of the unseen categories with the domain-specific features $\bm{S}_{A}$ and $\bm{S}_{B}$, we can generate samples of the unseen categories, which is formulated as:
\begin{equation}
	\bm{G}_{A}^{u} = \mathcal{G}([\bm{w}_{u};\bm{S}_{A}]),
	\bm{G}_{B}^{u} = \mathcal{G}([\bm{w}_{u};\bm{S}_{B}]) \end{equation}
where $\bm{G}_{A}^{u}\in\mathbb{R}^D$ and $\bm{G}_{B}^{u}\in\mathbb{R}^D$ are the generated samples of the unseen categories. Finally, we formulate the adversarial loss by minimizing the distance from the generated samples ($\bm{G}_{A}^{u}$, $\bm{G}_{B}^{u}$) to the initial features ($\bm{U}_A$, $\bm{U}_B$).

\subsection{Network Training Strategy}

\textbf{Loss functions.} In the semantic alignment module, the triple loss $\mathcal{L}_{tri}$ is used to enhance the discrimination of domain-invariant features, which is formulated as:
\begin{equation}
	\mathcal{L}_{tri} = max\{\|\bm{I}_{B}-\bm{I}_{A}\|_2 - \|\bm{I}_{B}-\bm{I}_{A} \|_2 + \eta, 0\}
\end{equation}
where the $\eta > 0$ is the margin parameter. In addition, we use the semantic loss $\mathcal{L}_{sem}$ to align the domain-invariant features with corresponding word embedding. Cosine similarity is used to measure the similarity between the domain-invariant features and word embeddings, $\mathcal{L}_{sem}$ can be formulated as:
\begin{equation}
	\begin{aligned}
		\mathcal{L}_{sem} = sim({E}_{A},\bm{w}_A) + sim({E}_{B},\bm{w}_A)  + sim({E}_{C},\bm{w}_C)
	\end{aligned}
\end{equation}
where the $sim(\cdot,\cdot)$ = $1- cos(\cdot,\cdot)$, $\bm{w}_A$ is the word embeddings of anchor sketch $\bm B$ and positive 3D shape $\bm A$, and $\bm{w}_C$ is the word embedding of negative 3D shape $\bm C$.

In the domain combination module, the reconstruction loss $\mathcal{L}_{rec}$ is formulated as:
\begin{equation}
	\begin{aligned}
		\mathcal{L}_{rec} = \|\bm{R}_{B}-\bm{F}_{B}\|_1 + \|\bm{R}_{A}-\bm{F}_{A}\|_1 + \|\bm{R}_{C}-\bm{F}_{C}\|_1
	\end{aligned}
\end{equation}
Note that we also consider the negative samples in the reconstruction loss. The adversarial loss $\mathcal{L}_{GAN}$ is formulated as:
\begin{equation}
	\begin{aligned}
		\mathcal{L}_{GAN} = \mathbb{E}( \log(D_{A}(\bm{F}_{A})) + \log(1 - D_{A}(\bm{G}_{A})) \\ + \mathbb{E}( \log(D_{B}(\bm{F}_{B})) + \log(1 - D_{B}(\bm{G}_{B}))
	\end{aligned}
\end{equation}

Furthermore, to mitigate the domain shift problem, we formulate the loss function $\mathcal{L}_{uGAN}$ to train our DD-GAN, which is computed as:
\begin{equation}
	\begin{aligned}
		\mathcal{L}_{uGAN} = \mathbb{E}( \log(D_{A}(\bm{U}_{B})) + \log(1 - D_{A}(\bm{G}_{A}^{u})) \\ + \mathbb{E}( \log(D_{B}(\bm{U}_{A})) + \log(1 - D_{B}(\bm{G}_{B}^{u}))
	\end{aligned}
\end{equation}
where the $\bm{U}_{B}$ and $\bm{U}_{A}$ are features of unlabeled sketches and 3D shapes extracted by ShapeCNN and SketchCNN respectively.

We finally train our model with the following loss:
\begin{equation}
	\begin{aligned}
		\mathcal{L}_{total}(E_A,E_s^A,E_B,E_s^B,G,D_A,D_B) = \\ \mathcal{L}_{tri} + \mathcal{L}_{sem} + \lambda_{rec}\mathcal{L}_{rec} + \mathcal{L}_{GAN} + \mathcal{L}_{uGAN}
	\end{aligned}
\end{equation}
where the $\lambda_{rec}$ is the weight to control the importance of cycle-consistency.

\textbf{Retraining and inference phase.} After DD-GAN generates unseen samples in the two domains of sketch and 3D shape, we use the real seen data and the generated unseen data to form a new training set to retrain our model.
The Unseen branch is removed and does not participate in retraining. 
At this time, the feature disentanglement in our model is mainly used to minimize the domain discrepancy between the two domains.
In the inference phase, we use the domain-invariant features of query sketches to search the domain-invariant features of 3D shapes.

\begin{figure}[ht]
	\centering
	\includegraphics[width=1.0\linewidth]{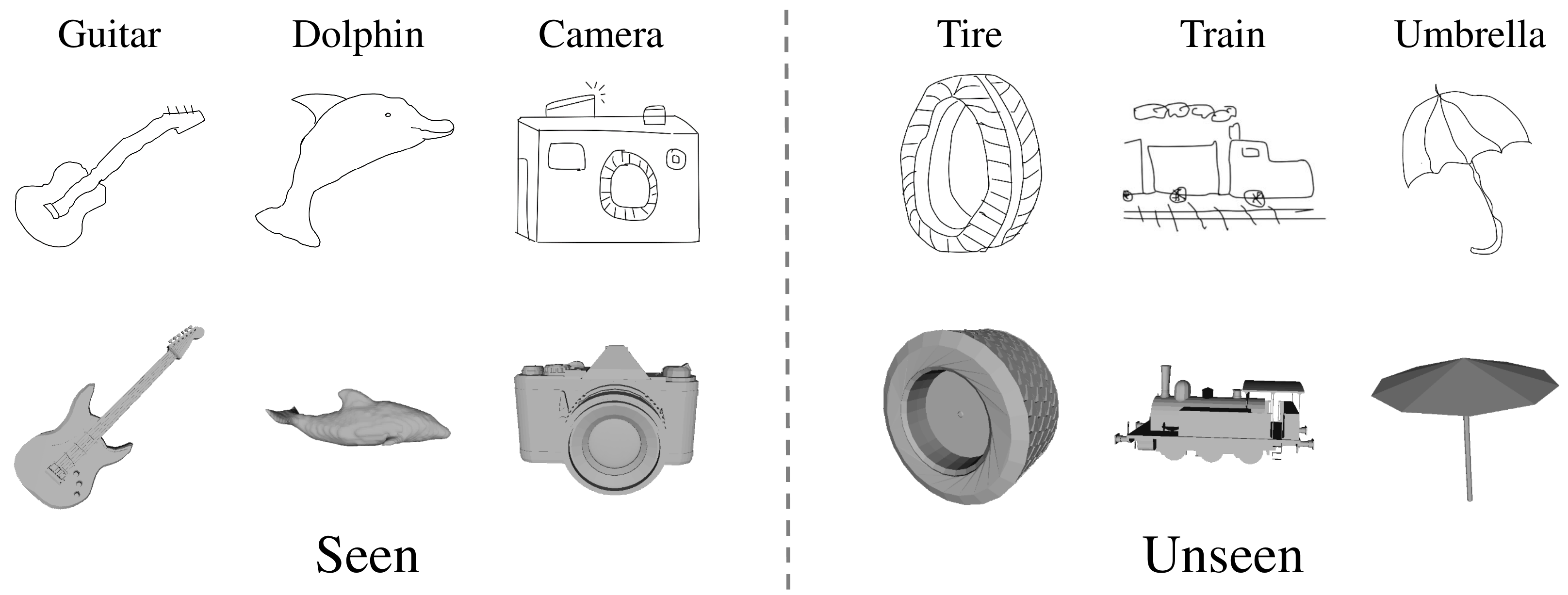}
	\caption{Examples of seen and unseen categories in the SHREC'14 dataset.}
	\label{fig:dataset_visual}
\end{figure}

\section{Experiment}

\subsection{Experimental Settings}
\textbf{Benchmark datasets and evaluation.} We evaluated our model on two widely used benchmarks, SHREC'13 \cite{li2013shrec} and SHREC'14 \cite{li2014shrec}, and compare with the state-of-the-art methods under the fair settings.

SHREC'13 is a benchmark for evaluate sketch-based 3D shape retrieval algorithms, which is created based on Princeton shape benchmark \cite{shilane2004princeton} and human sketch dataset.
There are a total of 90 categories in the dataset, including 7,200 hand-drawn sketches and 1,258 3D shapes.
Each class contains 80 sketches, but the number of 3D shapes is not equal.
The conventional sketch-based 3D retrieval methods divide each class of sketch into 50 for training and 30 for testing.
In this paper, the training set and test set are divided according to the category, 79 classes for training (seen) and 11 classes for testing (unseen).

SHREC'14 is larger than SHREC'13, which contains 13,680 hand-drawn sketches and 8,987 3D shapes.
Similar to SHREC'13, this dataset also contains 80 sketches for each category, 50 for training and 30 for testing.
We also re-divide the training set and the test set, 151 classes for training (seen) and 20 classes for testing (unseen).
As shown in Figure~\ref{fig:dataset_visual}, we visualize some examples of seen categories and unseen categories in the SHREC'14 dataset.

\textbf{Implementation details.} We implemented our DD-GAN using PyTorch. ResNet-50 \cite{he2016deep} pre-trained on ImageNet is used as backbone of SketchNet and ShapeNet. 
$E_i$ and $E_s$ have the same architecture with three fully-connected layers followed by two LeakyReLU, which output 300-D domain-invariant features and 300-D domain-specific features respectively.
We use the word text-based embedding model \cite{pennington2014glove} to extract 300-D word embeddings. 
The generator $G$ is a multi-layer perceptron containing three fully-connected layers generate the feature vectors of 2048-D.
The sketch domain and 3D shape domain discriminators $D_A$ and $D_B$ also share the same architecture with three fully-connected.
The margin $\eta$ in triplet loss is 20 and the weight $\lambda_{rec}$ of $\mathcal{L}_{rec}$ is 10 in this paper.
We adopt Adam to optimize our model with the learning rate of 1$e^{-5}$.

\begin{figure}[t]
	\centering
	\includegraphics[width=1.0\linewidth]{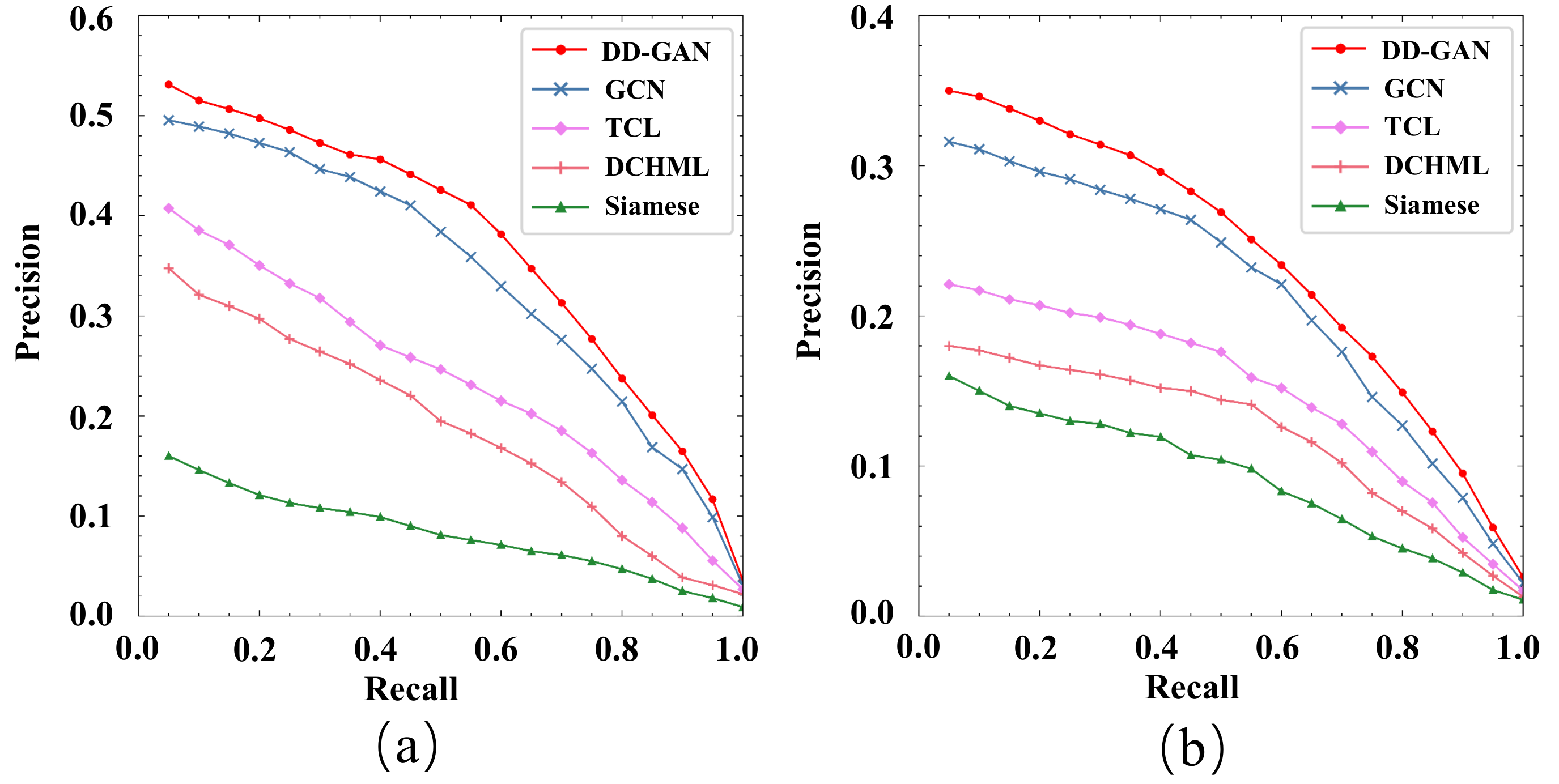}
	\vskip -10pt
	\caption{Performance comparison of precision-recall curve on the SHREC'13 dataset (a) and the SHREC'14 dataset (b).}
	\label{fig:PR13}
\end{figure}

\renewcommand{\arraystretch}{1.5}
\begin{table}[tb]
	\centering
	\fontsize{9}{8}\selectfont
	\begin{tabular}{lcccccc}
		\toprule
		Method&NN&FT&ST&E&DCG&mAP\\
		\midrule
		Siamese&0.137&0.114&0.203&0.162&0.404&0.171 \\
		DCHML& 0.318&0.304&0.421 &0.288&0.581&0.361  \\
		TCL&0.337 &0.357 &0.537 &0.278 &0.589 &0.426 \\
		CGN&0.512&0.458&0.647&0.347&0.673&0.515  \\
		\midrule
		Baseline&0.201&0.195&0.324&0.194&0.516&0.231 \\
		DD-GAN&\bf0.544&\bf0.484&\bf0.661&\bf0.364&\bf0.696&\bf0.551 \\
		\bottomrule		
	\end{tabular}
	\caption{Zero-shot retrieval results on the SHREC'13 dataset.}
	\label{tab:shrec13}

\end{table}

\subsection{Zero-Shot Sketch-Based 3D Shape Retrieval}
\textbf{Retrieval results on SHREC'13.} We compared our method with Siamese \cite{wang2015sketch}, DCHML \cite{dai2018deep}, TCL \cite{he2018triplet} and CGN\cite{dai2020cross} on the SHREC'13 dataset.
And we utilize the following widely-adopt metrics to evaluate our proposed method: nearest neighbor (NN), first tier (FT), second tier (ST), E-measure (E), discounted cumulated gain (DCG) and mean average precision (mAP).
For a fair comparison, we use the same backbone (ResNet-50) to extract the features of sketches and 3D shapes.
In addition, we train all the above methods under our zero-shot data division to be the same as our DD-GAN.
As we can see in Figure \ref{fig:PR13} (a), we apply precision-recall curve to compare our method with others, and our proposed method outperforms these methods.
The comparison results are also shown in Table \ref{tab:shrec13}. 
We use the model as our baseline, which has an encoder with the same architecture as $E_i$ and trained with the triple loss $\mathcal{L}_{tri}$.
It can be seen that the performance of the proposed DD-GAN method is significantly better than these methods.
Comparing with other methods, our model learns the knowledge from seen to unseen.
In addition, the generated high-quality unseen samples also let the model learn the distribution of real unseen samples.

\begin{figure}[t]
	\centering
	\includegraphics[width=1.0\linewidth]{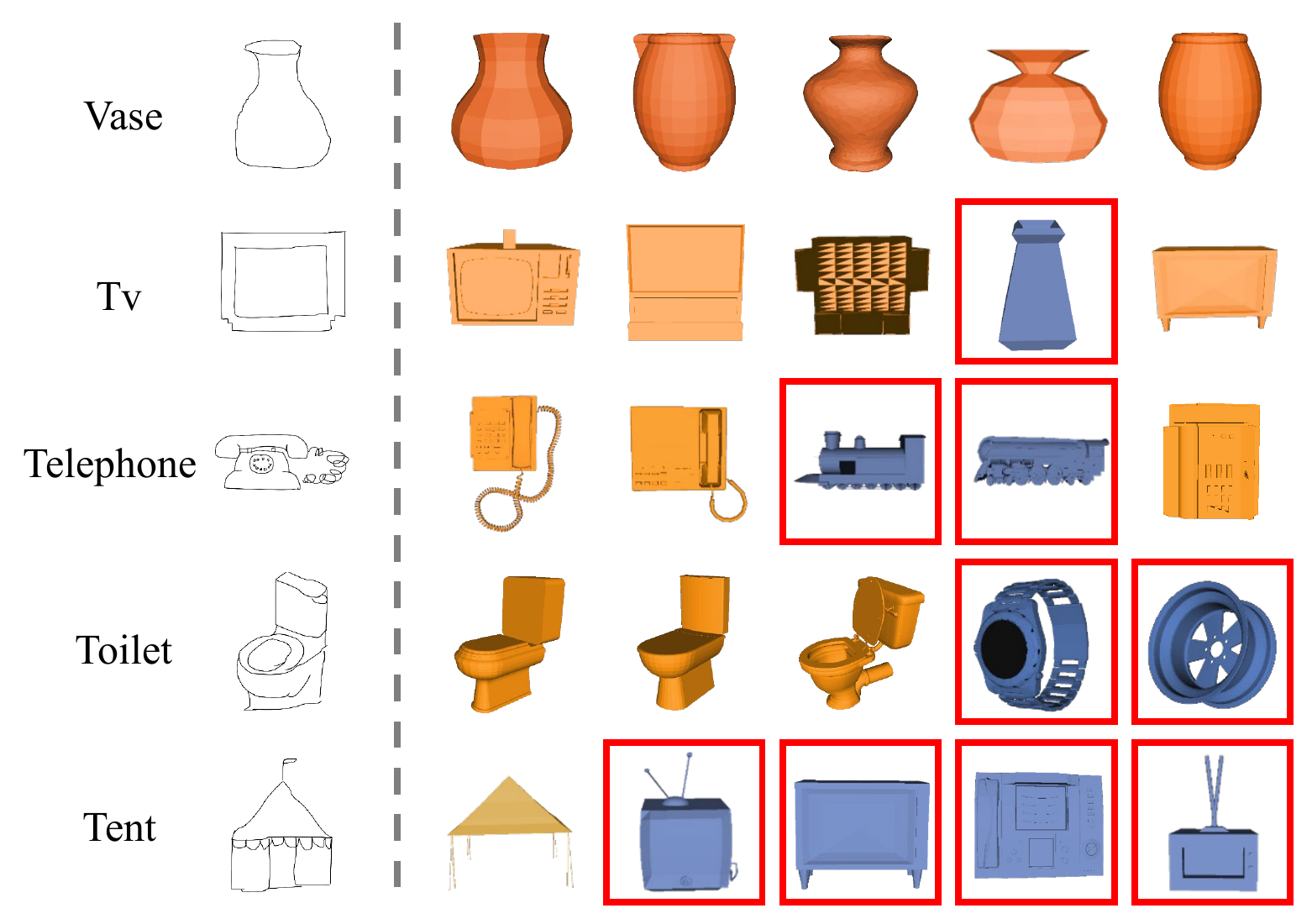}
	\vskip -10pt
	\caption{Top-5 zero-shot sketch-based 3D shape retrieval results obtained by our DD-GAN on the SHREC'13 dataset (top two rows) and the SHREC'14 dataset (next three rows). The failure cases are marked by red rectangles.
	}
	\label{fig:retrieval_visual}
\end{figure}

\renewcommand{\arraystretch}{1.5}
\begin{table}[htb]
	\centering
	\fontsize{9}{8}\selectfont
	\begin{tabular}{lcccccc}
		\toprule
		Method&NN&FT&ST&E&DCG&mAP\\
		\midrule
		Siamese&0.097&0.102&0.113&0.052&0.314&0.108 \\
		DCHML&0.157&0.134&0.145&0.084&0.379&0.187  \\
		TCL&0.279&0.257&0.153&0.125&0.459&0.237 \\
		CGN&0.401&0.324&0.429&0.178&0.571&0.332  \\
		\midrule
		Baseline&0.128&0.109&0.127&0.063&0.335&0.124 \\
		DD-GAN&\bf0.425&\bf0.354&\bf0.462&\bf0.196&\bf0.592&\bf0.371 \\
		\bottomrule
	\end{tabular}
	\caption{Zero-shot retrieval results on the SHREC'14 dataset.}
	\label{tab:shrec14}
\end{table}

\begin{figure*}[t]
	\centering
	\includegraphics[width=0.9\linewidth]{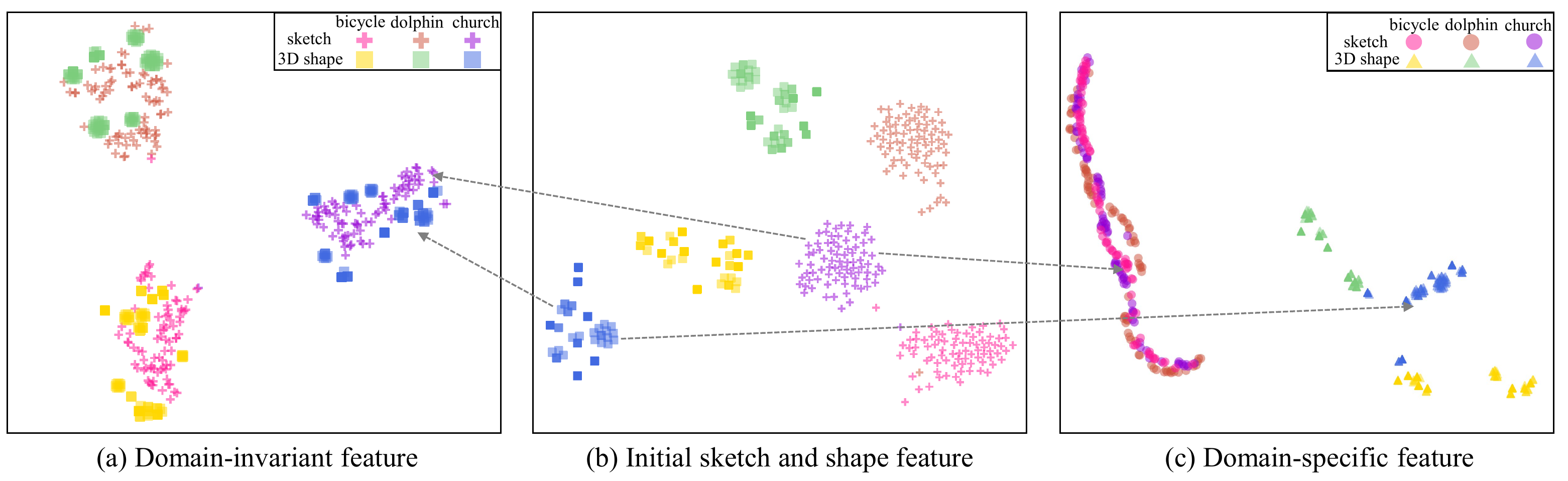}
	\caption{t-SNE Visualization of different features. Our domain disentangling module decomposes the initial sketch features and 3D shape features (b) into domain-invariant features (a) and domain-specific features (c). Compared with the initial sketch features and 3D shape features, domain-invariant features of the same class are easier to group together.
	}
	\label{fig:feature visualization}
\end{figure*}

\textbf{Retrieval results on SHREC'14.} We also evaluated our method on the SHREC'14 dataset and compared with Siamese \cite{wang2015sketch}, DCHML \cite{dai2018deep}, TCL \cite{he2018triplet} and CGN\cite{dai2020cross}.
The comparison results of NN, FT, ST, E, DCG and mAP are shown in Table \ref{tab:shrec14}. 
And the precision-recall curve on SHREC'14 is shown in Figure \ref{fig:PR13} (b).
Since the SHREC'14 dataset is larger than SHREC'13 and there are more categories, the retrieval task is also more difficult.
The retrieval performance of all methods is greatly reduced compared to the performance on SHREC'13 dataset, but our method is still the best.

\textbf{Qualitative results.} As shown in Figure \ref{fig:retrieval_visual}, we also visualize some successful and failed retrieval results of our DD-GAN on SHREC’13 and SHREC'14. For example, sketch query of ``Telephone'' retrieval some 3D shape of ``Train'' probably because the telephone receiver and train are both rectangular. For ``Toilet'' sketch, our model retrieves the ``Wristwatch'' and ``Wheel'' of 3D shape. Maybe the model captures that they all have circles. The ``Tent'' sketch and the ``TV'' 3D shape are also similar due to the rectangular structure. These wrongly retrieved 3D shapes usually have similar visual or semantics to the query sketches.

\renewcommand{\arraystretch}{1.5}
\begin{table}[htb]
	\centering
	\fontsize{8}{8}\selectfont
	\begin{tabular}{lcccccc}
		\toprule
		Method&NN&FT&ST&E&DCG&mAP\\
		\midrule
		Baseline&0.201&0.195&0.324&0.194&0.516&0.231 \\
		Baseline+SA& 0.275&0.224&0.362 &0.241&0.542&0.254  \\
		DD-GAN (O/T)&0.522 &0.464 &0.649 &0.351 &0.682 &0.523 \\
		DD-GAN (W/T)&\bf0.544&\bf0.484&\bf0.661&\bf0.364&\bf0.696&\bf0.551 \\
		\bottomrule
	\end{tabular}
	\caption{The performance on SHREC'13 dataset.}
	\label{tab:ablution}
\end{table}

\subsection{Ablation Study}

\textbf{Semantic alignment.} We conduct the experiments on the SHREC'13 dataset to verify the effectiveness of our proposed model.
It is worth noting that for the convenience of experiments, we use initial features extracted by SketchCNN and ShapeCNN as input instead of sketches and 3D shapes in subsequent experiments. Here we add semantic loss $\mathcal{L}_{sem}$ to the baseline to demonstrate the effectiveness of the semantic alignment module. 
As shown in Table \ref{tab:ablution}, semantic alignment module (Baseline $+$ SA) improves the retrieval performance of the model on the unseen category.
This can prove that our semantic alignment module can transfer knowledge from seen classes to unseen classes.

\textbf{Domain disentangling.} As shown in Figure \ref{fig:feature visualization}, we use t-SNE \cite{van2008visualizing} to visualize features of sketches and 3D shapes from different three categories.
In Figure \ref{fig:feature visualization} (b), we use our baseline with semantic loss $\mathcal{L}_{sem}$ to extract initial features of sketches and 3D shapes (Baseline $+$ SA).
We can see that although the features of sketches and 3D shapes are discriminative, the same categories of sketch features and 3D shape features still have a domain gap.
To prove the effectiveness of domain disentangling, we added domain disentangling and domain combination to the model (DD-GAN(O/T)). 
The reason why we add domain combination here is that it ensures that domain-specific features and domain-invariant features can be decoupled.
As shown in Figure \ref{fig:feature visualization} (a) and Figure \ref{fig:feature visualization} (c), after disentangling domain-specific features, the domain-invariant features of the same class in the two domains can be effectively narrowed in the common feature space.
For example, in Figure \ref{fig:feature visualization} (b), although the 3D shape features and sketch features of ``Church'' belong to the same category, they are still away from each other.
After feature disentanglement, the domain-invariant features of ``Church'' in the two domains are close together, as shown in Figure \ref{fig:feature visualization} (a).
In Table \ref{tab:ablution}, the quantitative evaluation of the DD-GAN(O/T) is also much better than (Baseline $+$ SA).

\begin{figure}[htb]
	\centering
	\includegraphics[width=1.0\linewidth]{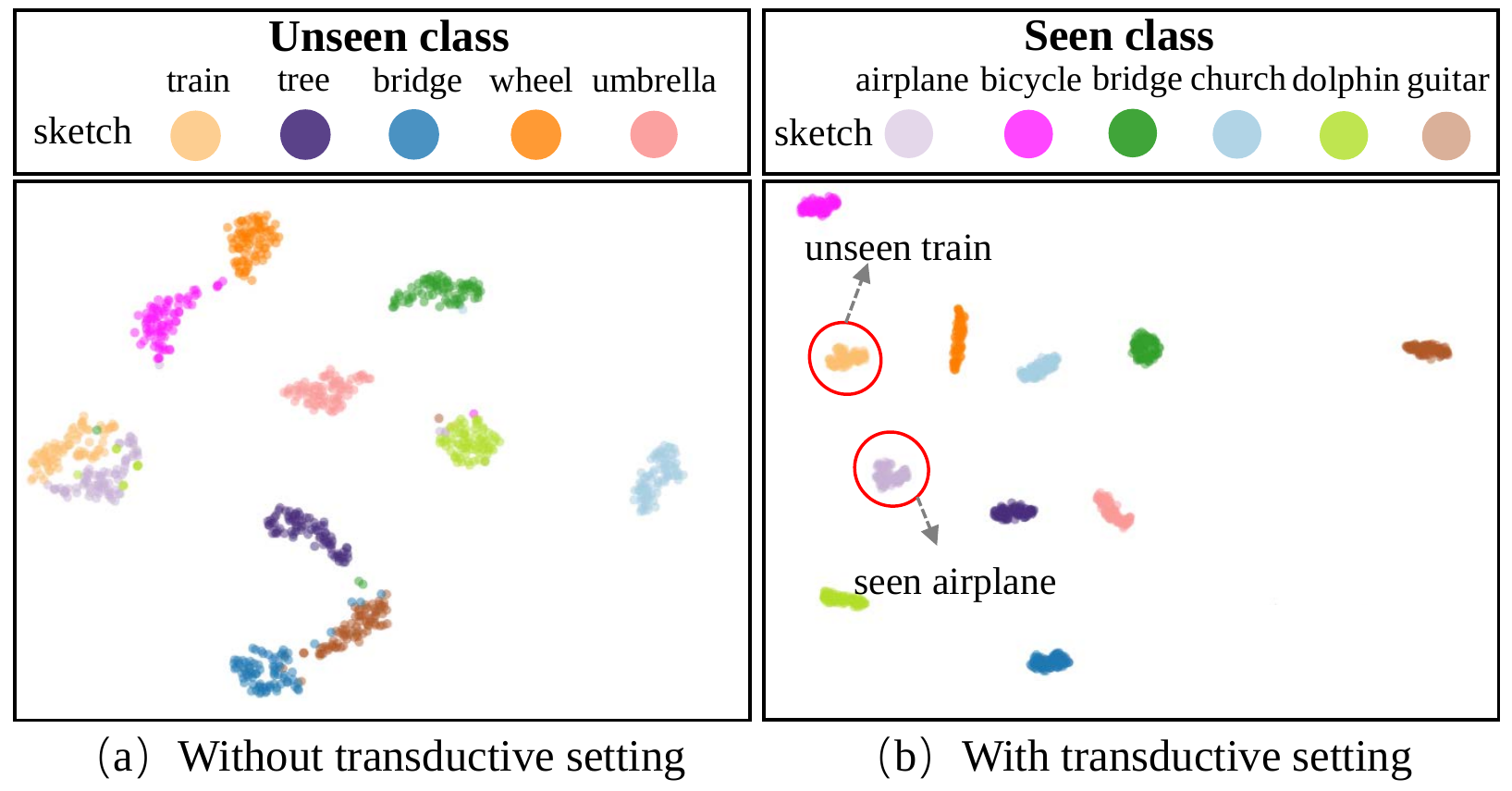}
	
	\caption{t-SNE Visualization of different features. Comparing (a) and (b), after performing the transductive setting, the generated unseen samples can be better distinguished from the real visible samples.
	}
	\label{fig:feature visualization2}
\end{figure}

\textbf{Transductive setting.} We compare our full model DD-GAN (W/T) to its variants without transductive setting (DD-GAN(O/T)).
We first generate unseen samples of sketch domain through the two models, which combine the unseen word embedding with the domain-specific feature of seen class sketch.
Then we apply t-SNE to visualize the generated samples together with real samples of seen classes in Figure \ref{fig:feature visualization2}.
As shown in Figure \ref{fig:feature visualization2} (a), the DD-GAN(O/T) has not seen the real unseen samples, which make the generated unseen samples are similar to the real seen categories.
For example, the generated unseen samples of the ``Train'' category are close to the real seen class ``Airplane".
This will reduce the quality of the generated unseen samples, resulting in little improvement in the model retraining effect.
In Figure \ref{fig:feature visualization2} (b), the unseen samples generated by DD-GAN(W/T) are obviously different from the real seen categories.
In addition, as shown in Table \ref{tab:ablution}, the retrieval performance of DD-GAN(W/T) is better than DD-GAN(O/T).
This proves that we introduced unlabeled data in an unsupervised way, which effectively avoided the domain shift problem and improved the quality of the generated samples.

\section{Conclusion}
In this paper, we are the first to explore zero-shot sketch-based 3D shape retrieval. In order to make the model effectively retrieve unseen categories, we propose a domain disentangled generative adversarial network (DD-GAN). Our model can not only reduce the inter-domain difference between sketch and 3D shape, but also minimize the domain discrepancy between seen categories and unseen categories. Extensive experiments on the SHREC'13 dataset and SHREC'14 dataset can demonstrate the effectiveness of the proposed method for zero-shot sketch-based 3D retrieval. In future work, we will consider exploring a more effective way to generate high-quality unseen samples in the zero-shot sketch-based 3D shape retrieval task.

\section{Acknowledgments}
This work was supported by the National Science Foundation of China (Grant Nos. 61876084, 61876083, 62176124), and the Postgraduate Research \& Practice Innovation Program of Jiangsu Province (No. KYCX21 0302).

\bigskip

\bibliography{aaai22}
\end{document}